\documentclass{article}
\usepackage{spconf,amsmath,graphicx,hyperref}
\usepackage{pifont}
\usepackage{booktabs}
\usepackage{multirow}
\usepackage[table]{xcolor} % 重要：让颜色作用于表格
\usepackage{soul}
\usepackage{xcolor}

\title{TextlessRAG: End-to-End Visual Document RAG by Speech Without Text\thanks{\copyright  2025 IEEE. Personal use of this material is permitted. Permission from IEEE must be obtained for all other uses, in any current or future media, including reprinting/republishing this material for advertising or promotional purposes, creating new collective works, for resale or redistribution to servers or lists, or reuse of any copyrighted component of this work in other works.}}
\twoauthors
 {Peijin Xie, Shun Qian, Bingquan Liu}
	{Harbin Institute of Technology\\
	ITNLP Lab\\
	Harbin, China}
 {Dexin Wang, Lin Sun, Xiangzheng Zhang}
	{Qihoo 360 Technology\\
	ZhiNao AI Lab\\
	Beijin, China}

\begin{document}

%\ninept
%
\maketitle
\begin{abstract}

% 文档图片包含了大量的知识，而语音提问的便携性有助于使用场景的扩展，但是目前却没有基于语音为query的文档知识库问答研究。
% 我们首次提出了一个去文本化的端到端语音知识库问答方案TextlessRAG, 该方案直接理解语音然后检索文档图片知识库最后做问答。
% 过程中没有用ASR语音转文本也没有使用OCR图像转文本。
% 实验表明我们简洁的端到端方案在效率和准确率上体现出极大优势。
% 其次我们搭建了一个data engine 构造了首个语音文档rag数据集。
% 分别包含中文和英文的语音提问与文档知识内容。
% 数据集与pipeline将开源在url
% 摘要应包含大约100到150个单词。
Document images encapsulate a wealth of knowledge, while the portability of spoken queries enables broader and flexible application scenarios. 
Yet, no prior work has explored knowledge base question answering over visual document images with queries provided directly in speech.
We propose TextlessRAG, the first end-to-end framework for speech-based question answering over large-scale document images. 
Unlike prior methods, TextlessRAG eliminates ASR, TTS and OCR, directly interpreting speech, retrieving relevant visual knowledge, and generating answers in a fully textless pipeline.
To further boost performance, we integrate a layout-aware reranking mechanism to refine retrieval.
Experiments demonstrate substantial improvements in both efficiency and accuracy. 
To advance research in this direction, we also release the first bilingual speech–document RAG dataset, featuring Chinese and English voice queries paired with multimodal document content.
Both the dataset and our pipeline will be made available at repository:\url{https://github.com/xiepeijinhit-hue/textlessrag}
% Experimental results show that this streamlined, end-to-end design delivers substantial gains in both efficiency and accuracy. 
% In addition, we develop a data engine to construct the first speech–document RAG dataset, containing voice queries in both Chinese and English paired with corresponding document knowledge content. 

% The abstract should appear at the top of the left-hand column of text, about
% 0.5 inch (12 mm) below the title area and no more than 3.125 inches (80 mm) in
% length.  Leave a 0.5 inch (12 mm) space between the end of the abstract and the
% beginning of the main text.  The abstract should contain about 100 to 150
% words, and should be identical to the abstract text submitted electronically
% along with the paper cover sheet.  All manuscripts must be in English, printed
% in black ink.
\end{abstract}
\section{Introduction}
\label{sec:intro}
\begin{figure}[h] % 这里控制图片位置：h=here, t=top, b=bottom, p=page
    \centering
    \includegraphics[width=0.5\textwidth]{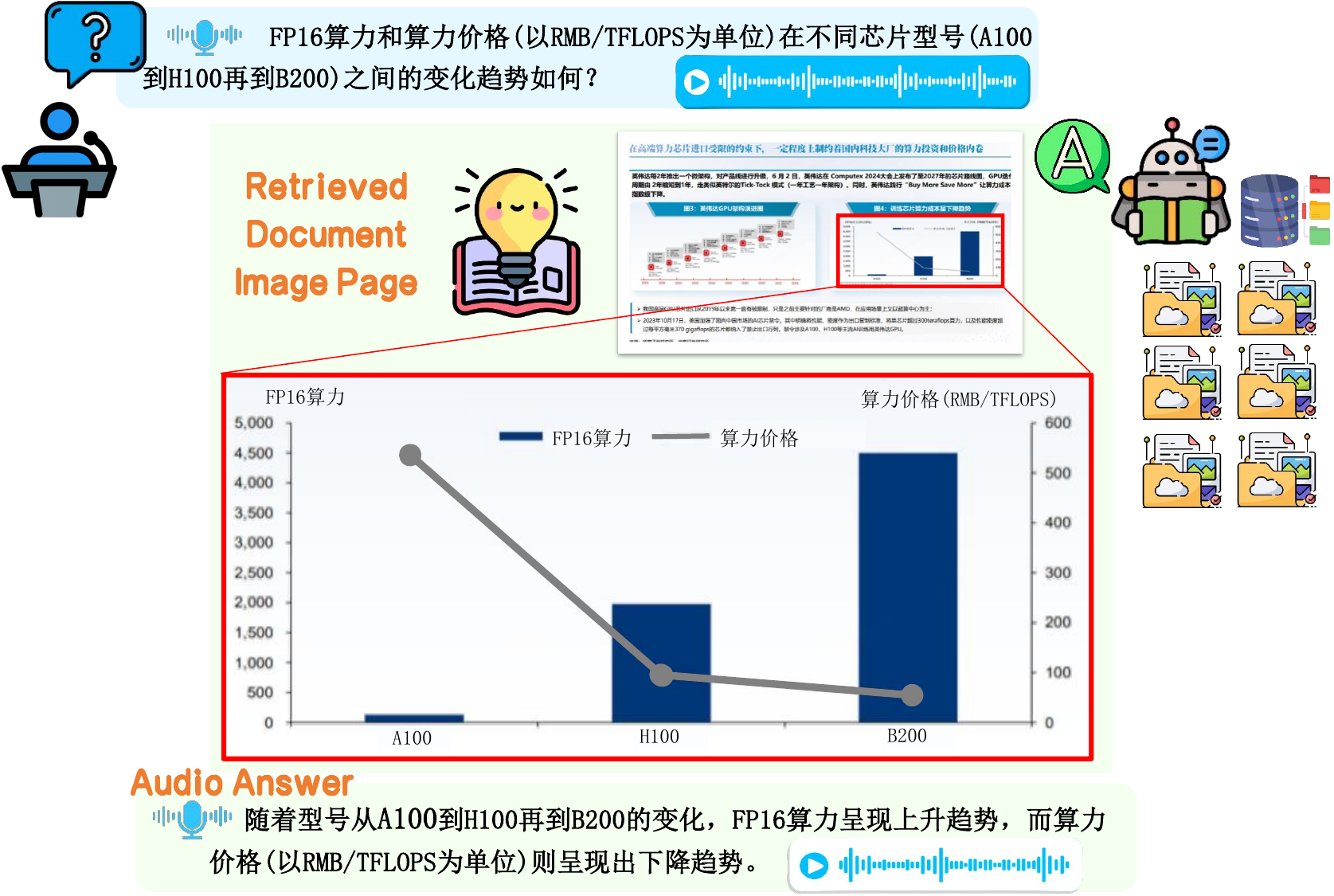} % 图片路径和宽度
    \caption{A Chinese example of TextlessRAG.}
    \label{fig:example} % 标签，用于引用
\end{figure}
% User inputs queries via speech. Model performs retrieval on a knowledge base  to obtain relevant images and regions to generates answer  in audio.
% 文档里包含了密集的知识-》 视觉文档模型发展， 知识库问答

% 日常使用的时候用户需要键入问题还需要找出对应文档，手里既需要有现成文档有需要有键入键盘

% 介绍文档理解现状：,文档中存贮着丰富的知识，提出现有问题：
% 1.在文书文档处理领域，视觉文档理解发展好,随着
% 2.rag 减少幻觉
% 3.问题： 缺少语音提问数据集与rag 方案，语音提问的便携性可以扩展使用场景，而不单单只是文档处理

% VLM 的发展极其迅速，随着动态分辨率视觉编码的提出
% 近期VLM在文字密集型图片上的理解能力得到了很大的飞跃
% 模型在直接的视觉文档理解上取得了很大的进步，
% 在模型取得更高正确率的同时模型也省去了以往文档解析的步骤，使得整个理解过程变得更加高效

% 现有研究也是慢慢转向面向更长的更复杂的文档理解任务，包括长文档问答以及知识库问答。
% 但是目前却从来没有工作探索过在文档图片上基于口语语音提问的工作。
% 可以想象基于语音的知识库问答是非常便捷的，因为它既不需要键盘键入提问文本，也不需要用户上传待提问文档图片。
% 人们直接口述提问就可以得到答案，这样的方式将会极大扩展了多模态大语言模型的使用场景。

% 于是我们首先构造了首个基于语音输入的文档视觉rag pipeline xxx.
% xxx 实现了首先直接使用语音查询相关的视觉文档知识信息，检索出相关视觉内容，
% 然后多模态大语言模型直接理解语音内容根据视觉知识回答问题。
% 整个流程没有文档解析OCR过程，没有ASR识别，也没有TTS生成。
% 全程无文本化保证了整个过程的高效率以及对图表表格等多模态内容的准确理解。
% 期间我们还使用了layout rerank方法进一步精炼了提取的视觉知识提高了模型的表现.

% 此外我们还首次建立了首个基于语音的视觉知识库rag  bench 数据集xxx。
% xxx 包括六个主流英文视觉文档理解以及rag的语音扩展和自己人工标注的中文数据集 yyy。
% 在数据集上的测试我们发现，xxx这种基于语音的视觉RAG方式无论是回答质量还是在回答速率上都有极大的优势。

% \cite{Qwen2.5-Omni},\cite{TM-VQA}
The development of Vision-Language Models (VLMs) has been remarkably rapid\cite{Qwen2.5-VL, Qwen2-VL, zhu2025internvl3exploringadvancedtraining,Qwen2.5-Omni}. 
With the introduction of dynamic-resolution visual encoders, recent VLMs have made significant advances in understanding text-rich images\cite{multimodalLLM}
These models not only achieve higher accuracy in direct visual document understanding but also eliminate the need for traditional document parsing steps, thereby making the entire process more efficient.

Current research is increasingly shifting towards more complex long-document understanding tasks, including long-form question answering\cite{mmlongbenchdoc, SlideVQA2023} and knowledge base QA from RAG\cite{visrag, vdocrag, vidoragvidoseek}. 
However, no prior work has explored speech-based querying over document images; existing studies on spoken input have been limited to extending visual question answering on natural images\cite{TM-VQA, spokenvqa}.
Spoken knowledge base QA is particularly appealing: it requires neither typing queries on a keyboard nor uploading document images. 
Instead, users can obtain answers simply by asking questions verbally—an interaction mode that can greatly expand the application scenarios of multimodal large language models.

% 用户直接口述问题，模型首先从知识库里检索出符合查询的页面，然后使用layout切分出相关的回答依据块（如图表表格自然图片或者文本块等），最后根据细粒度的依据块模型回答问题直接输出语音以及文本答案内容。
To this end, we propose \textbf{TextlessRAG}, the first speech-based document visual RAG pipeline. As illustrated in Figure \ref{fig:example}, the user poses a question through spoken input. The model then retrieves candidate pages from the knowledge base, followed by layout-aware decomposition of these pages into fine-grained evidence units (e.g., charts, tables, natural images, and textual segments). Based on the selected evidence, the model generates an answer and delivers the response in both speech and text.
Notably, the entire pipeline is fully de-textualized—operating without OCR, ASR, or TTS—thereby ensuring both efficiency and robust comprehension of multimodal knowledge, particularly for structured content such as tables and charts.
% In addition, we incorporate a layout reranking strategy to refine retrieved visual knowledge, further enhancing model performance.

We further construct the first speech-based visual document RAG benchmark (\textbf{SV-DOC}), which comprises speech-augmented extensions of existing English visual document QA datasets\cite{chartqa, infographicvqa, dude, SlideVQA2023, mmlongbenchdoc} and RAGBench\cite{vidoragvidoseek}, along with a manually annotated Chinese Document RAG dataset (CDR). Experimental results demonstrate that our speech-based visual RAG approach delivers significant improvements in both answer quality and response latency.

% We also construct the first Speech-based Visual Document RAG benchmark, SV-DOC, which includes speech-augmented extensions of the English visual document QA datasets\cite{chartqa, infographicvqa, dude, SlideVQA2023, mmlongbenchdoc} and RAG bench\cite{vidoragvidoseek}, as well as a manually annotated Chinese Document RAG dataset, CDR. 
% Experimental results show that our proposed speech-based visual RAG approach achieves substantial advantages in both answer quality and response speed.

% 引入TextlessRAG，解决问题，好处：
% 1.使用场景广泛
% 2.端到端：无中间OCR，ASR 过程
%     低时延
%     高精度，
% 3.使用block reranking 优化检索

% 搭建首个RAG Framwork
% 检索用....问答用...

% 构造首个数据集：
% 现有英文数据集扩展
% 中文数据构建

% 实验结果，
% 主要结果：检索结果，问答结果，仍存在很大挑战，瓶颈在于检索
% 正确率提升： 与纯文本方案对比，与ASR加纯文本方案对比
% 效率提升

% The main contributions are summarized as follows:
% \begin{itemize}
%     \item We propose TextlessRAG, the first speech-based, end-to-end visual document RAG pipeline that operates without ASR, OCR, or TTS.
%     \item  We construct SV-DOC, the first speech-based visual document RAG benchmark, featuring bilingual data in English and Chinese.
%     \item The text-free workflow design, combined with the layout reranking strategy, delivers significant gains in both efficiency and accuracy throughout the RAG process.
% \end{itemize}

The main contributions of this work are summarized as follows:
\begin{itemize}
    \item We introduce \textbf{TextlessRAG}, the first end-to-end speech-based visual document RAG pipeline, which operates entirely without ASR, OCR, or TTS. 
    \item We present \textbf{SV-DOC}, the first benchmark for speech-based visual document RAG, featuring bilingual datasets in English and Chinese. 
    \item We design a fully text-free workflow augmented with a layout-aware reranking strategy, achieving substantial improvements in both efficiency and accuracy across the RAG process. 
\end{itemize}

% 主要贡献
% 1.首个pipline
% 2.data engine ，首个data
% 3.带来新的挑战

\section{Method}
\label{sec:method}

\begin{figure}[tp] % 这里控制图片位置：h=here, t=top, b=bottom, p=page
    \centering
    \includegraphics[width=0.4\textwidth]{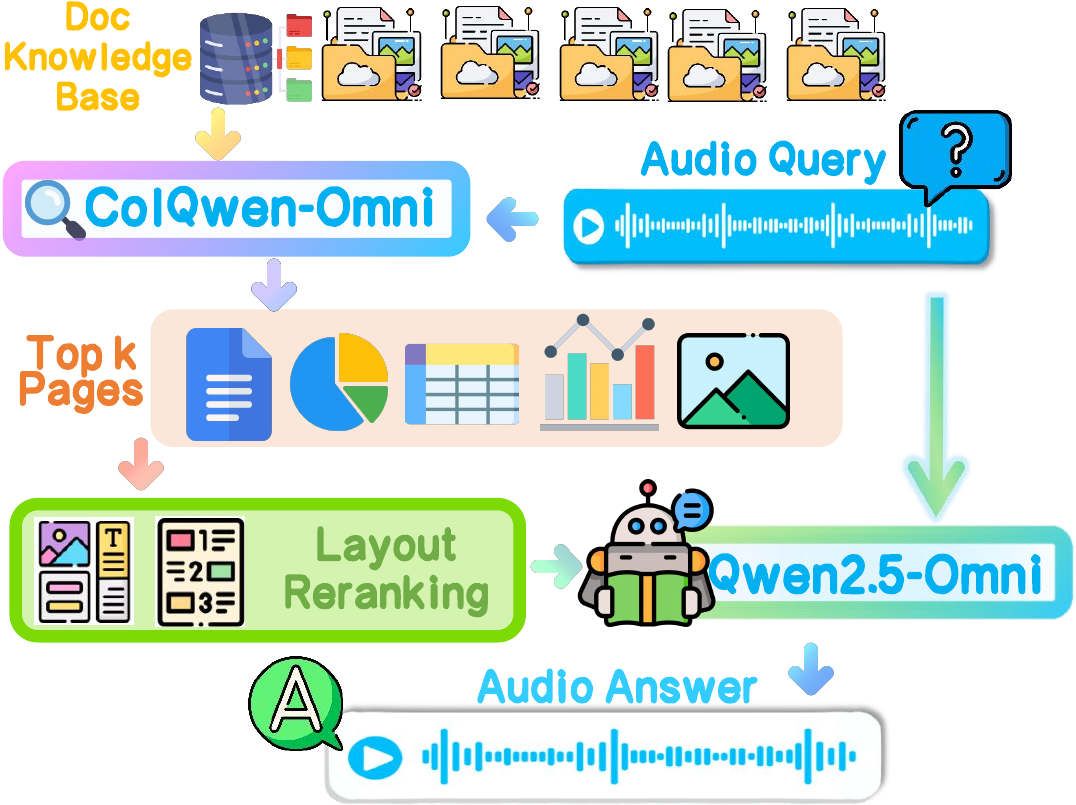} % 图片路径和宽度
    \caption{Pipeline of TextlessRAG. }
    \label{fig:pipeline} % 标签，用于引用
\end{figure}

As shown in Figure \ref{fig:pipeline}, TextLessRAG is designed to be both concise and efficient focusing specifically on the retrieval and QA generation components.
\textbf{On Retrieval side}, we employ ColQwen-Omni\footnote{\url{https://huggingface.co/vidore/colqwen-omni-v0.1}} as a retrieve encoder $\mathrm{Enc}(\cdot)$ to encode both document image pages from document knowedge base $\mathcal{I}=\{P_1, P_2...P_n\}$and spoken query $q$ to embeddings as $E=\{e_1, e_2...e_n\}$ and $e_q$. 
Then compute the MaxSim Score like ColBert\footnote{\url{https://github.com/stanford-futuredata/ColBERT}} to rank the top-k candidate pages $T_k$:
\begin{equation}
E = \{ e_i = \mathrm{Enc}(I_i) \mid I_i \in \mathcal{I} \}, \quad 
e_q = \mathrm{Enc}(q)
\end{equation}
\begin{equation}
s_i = Score(P_i, q)=\mathrm{MaxSim}\big({e_q}, e_i\big),
\quad e_i \in E
\end{equation}
\begin{equation}
\mathcal{T}_k = \operatorname{TopK}_{P_i \in \mathcal{I}} 
\Big( Score(P_i, q) \Big) = [P_{t_1},P_{t_2}\dots P_{t_k}]
\end{equation}

\textbf{On Generation side}, we employ Qwen2.5-Omni\footnote{\url{https://huggingface.co/Qwen/Qwen2.5-Omni-7B}} as a Generator $Gen(\cdot)$, which directly takes the spoken query $q$ together with the top-k re-ranked document images $\mathcal{T}_k'$ as input, and ultimately produces a spoken answer $\mathcal{A}ns$:
\begin{equation}
\mathcal{A}ns = Gen(\mathcal{T}_k',q), \quad \mathcal{T}_k'=[P_{t_1}',P_{t_2}'\dots P_{t_k}']
\end{equation}
% To optimize the answer generation, we reorganized the image inputs to refine representations that are more relevant to the given question:

We lay out each top-k page $P_t$ into content blocks of chart, table, text and natural image sets by DocLayout-Yolo\footnote{\url{https://github.com/opendatalab/DocLayout-YOLO}}:
\begin{equation}
\mathcal{C}ontent = \{\text{chart}, \text{table}, \text{text}, \text{image}\}
\end{equation}
\begin{equation}
YOLO(P_t) = \bigcup_{c \in \mathcal{C}ontent} Block^{c}(P_t), 
\quad P_t \in \mathcal{T}_k
\end{equation}
and refine them with threshold $\theta$ as a lower bound to guarantee the MaxSim Score with query $s_b$:
\begin{equation}
{Block}^{c}_{\theta} = 
\bigcup_{c \in \mathcal{C}ontent} 
\left\{ b \in Block^{c}(P_t) \;\middle|\; s_b \geq \theta \right\}
\end{equation}

Then, we resort the image inputs $P_t$ to refine representations $P_t'$ that are more relevant to the given question:
\begin{equation}
 P_t' \;=\; \operatorname{Sort}_{\downarrow Score}\big({Block}^{c}_{\theta}\big),
\end{equation}

Overall, the end-to-end process facilitates a fully text-free workflow: it requires neither ASR nor OCR and obviates the need for text-to-answer TTS, enabling direct spoken queries and responses. 
Furthermore, the layout segmentation and embedding of knowledge base document images are preprocessed in advance. 
Each query involves two retrieval stages, which not only improves the relevance of retrieved results but also substantially reduces computational overhead.
% 最终我们的pipeline可以表示为。。公式。。整个端到端过程没有xxx 没有xxx，知识库的图片和layout都事先编码好，整个端到端过程语音输入语音输出，直接给出回答。
% \subsection{Retrive}

\section{Data Engine}
\label{sec:data}
\begin{figure}[tbp] % 这里控制图片位置：h=here, t=top, b=bottom, p=page
    \centering
    \includegraphics[width=0.47\textwidth]{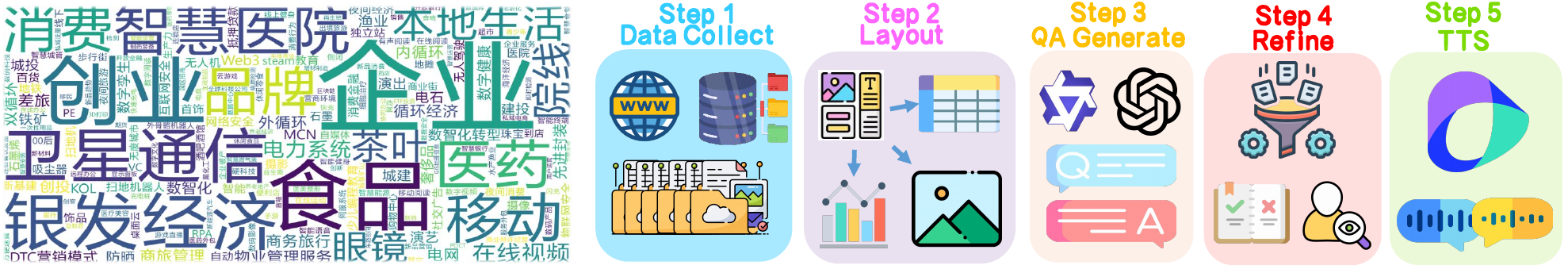} % 图片路径和宽度
    \caption{Word Cloud of domains of Chinese samples on the left and the 5 steps data engine on the right.}
    \label{fig:data} % 标签，用于引用
\end{figure}

We developed a data engine to construct the SV-DOC benchmark. As summarized in Table \ref{table:data}, SV-DOC comprises (i) extensions of existing English benchmarks augmented with TTS, and (ii) CDR, the first Chinese Document RAG dataset created from scratch. CDR is built following a five-step pipeline (illustrated on the right of Figure \ref{fig:data}), while the final two steps are additionally used to generate speech-augmented English samples.

Specifically, (1) we collect a diverse set of documents and reports across multiple domains and formats, including PDFs and PPTs (see the domain word cloud in Figure \ref{fig:data}, left). (2) Each page is segmented into fine-grained units—tables, charts, images, and text blocks—using DocLayout-YOLO. (3) These segments are processed by commercial vision–language agents via batch APIs to generate candidate QA pairs. (4) The QA pairs are refined through rule-based filtering and professional human annotation. (5) Finally, the text QA pairs are converted into speech using Doubao’s TTS API, leveraging over 200 professional voice types.\footnote{\url{https://www.volcengine.com/docs/6561/1257544}}. 

For the English datasets, we select refined, relevant visual document QA datasets—ChartQA\cite{chartqa}, DUDE\cite{dude}, InfoVQA\cite{infographicvqa}, and SlideVQA\cite{SlideVQA2023}(following VDocRAG\cite{vdocrag}), as well as MMLongBench-Doc (MMLong)\cite{mmlongbenchdoc} and Vidoseek\cite{vidoragvidoseek} — and apply steps 4 and 5 to augment them with audio queries.
Further details of the data construction process and representative examples are provided in the repository.

% (chartqa, dude, infovqa slidevqa)
% In the first step, we collect a large number of documents and reports from diverse domains on the Internet, in various formats including PDF and PPT. 
% The domain-specific word cloud is shown on the left side of Figure 2. In the second step, we use YOLO to segment the layout of each page, extracting fine-grained table, chart, image, and text blocks. 
% In the third step, the extracted segments are fed to multiple vision-language (VL) agents, invoking batch APIs to generate text-based question-answer pairs for each content block. In the fourth step, the generated text QA pairs are refined using a combination of rule-based methods and professional annotators. 
% Finally, in the fifth step, we employ Doubao’s TTS API to convert the text QA pairs into spoken form, using as many as 200 different voice types whenever possible.

\begin{table}[]
\centering
% \resizebox{\columnwidth}{!}{%
\begin{tabular}{ccccc}
\toprule
\textbf{Dataset}&\textbf{QA}&\textbf{Pool}&\textbf{Content}&\textbf{Domain} \\ \hline
Chartqa\cite{chartqa}         &150            &119          &C               &Academic        \\
Infovqa\cite{infographicvqa}         &1048           &300          &C/T/I         &Infographic     \\
Slidevqa\cite{SlideVQA2023}       &760             &657           &C/T/I/X          &Slide           \\
DUDE\cite{dude}            &496            &422          &C/T/X           &Open            \\
MMLong\cite{mmlongbenchdoc}  &1091            &5134      &C/T/I/X        &Open            \\   %135/5134  MMlongbench-doc
Vidoseek\cite{vidoragvidoseek}         &1142            &5349      &C/T/I          &Open            \\    % 290/5349 
CDR    &1260            &30583               &C/T/I/X        &Open            \\ \hline
Total            &5947            &42564               &C/T/I/X        &Open            \\ 
\bottomrule
\end{tabular}
% }
\caption{Data statistics of SV-Doc Bench.``QA" and ``Pool" calculate the amount of qa pairs and images of retrieval pool. The multimodal content of chart, tabel, natural image and text are abbreviated as ``C", ``T", ``I" and ``X".}
\label{table:data}
\end{table}

\section{Experiment \& Result}
\label{sec:exp}
\begin{table*}[htpb]
\centering
\resizebox{\textwidth}{!}{
\begin{tabular}{ccccccccccc}
\toprule
Model &Retriver &Query &Doc&ChartQA&DUDE&Infovqa&SlideVQA&MMLong&Vidoseek&CDR \\ \hline
BM25        &-            &T  &T  &54.8 &57.2 &50.2 &40.7 &18.5 &84.5 &54.9 \\
E5          &BERT         &T  &T  &74.9 &40.6 &42.5 &50.8 &23.4 &63.5 &62.6 \\
NV-Embed-v2 &Mistral      &T  &T  &75.3 &43.0  &56.5 &61.7 &38.7 &90.3 &69.3 \\  \hline
CLIP        &Scratch      &T  &I &54.6 &23.2 &29.7 &38.6 &17.3&35.8 &32.5 \\
DSE         &Phi3V        &T  &I &72.7 &55.5 &67.4 &73.0 &43.6 &89.4 &77.1 \\
VisRAG-Ret  &MiniCPM-V    &T  &I &87.2 &56.4 &71.9 &74.3 &53.1 &91.2 &80.9 \\
VDocRAG     &Phi3V        &T  &I &86.0 &57.7 &72.9 &77.3 &49.2 &92.8 &82.4 \\
ViDoRAG    &Colqwen2     &T  &I &\textbf{100}  &\textbf{96.5} &\textbf{97.8} &\textbf{96.9} &\textbf{67.0} &94.3 &\textbf{87.7} \\ \hline
TextLessRAG &Colqwen-Omni &A &I &99.3  &91.5 &91.6 &94.2 &66.5 &\textbf{95.4}  &87.4 \\ \bottomrule
\end{tabular}
}
\caption{Rertrive result. ``T", ``I", and ``A" represent the query and documents in the form of text, image, and audio, respectively.}
\label{table:retrive}

\end{table*}

% \cellcolor{yellow!30}

\begin{table*}[tpb]
\centering
\begin{tabular}{ccccccccc}
\toprule
Model             &Generator&ChartQA&DUDE&InfoVQA&SlideVQA&MMLong&Vidoseek&CDR \\ \hline
TextRAG           &\multirow{2}{*}{Phi3}        &28.0      &40.1 &40.5    &28.6     &6.9                 &29.8         &10.5              \\
TextRAG\dag       &    &36.6    &55.9 &45.6    &27.8     &13.1                 &31.7         &18.7              \\ \hline
VDocRAG           &\multirow{2}{*}{Phi3V}         &52.0      &48.5 &56.2    &48.0       &14.5                 &52.1         &22.3              \\
VDocRAG\dag       &    &74.0      &66.4 &64.6    &56.4     &21.7                 &63.8         &34.6              \\ \hline
ViDoRAG           &\multirow{2}{*}{Qwen2.5-VL7B}  &84.6        &\cellcolor{green!25}86.7      &79.1         &82.5          &\cellcolor{green!25}37.9                 &85.7     &\cellcolor{green!25}50.0               \\
ViDoRAG\dag       &    &84.6         &\cellcolor{yellow!30}87.4      &\cellcolor{yellow!30}82.6         &\cellcolor{yellow!30}84.2          &\cellcolor{yellow!30}47.3                 &86.4          &\cellcolor{yellow!30}70.1               \\\hline
TextLessRAG      &\multirow{3}{*}{Qwen-Omni}     &87.3         &78.5      &74.5         &79.7          &33.4                 &90.2          &43.5               \\ 
TextLessRAG*     &    &\cellcolor{green!25}87.3    &81.3     &\cellcolor{green!25}79.4        &\cellcolor{green!25}82.6         &36.7     &\cellcolor{green!25}93.4         &47.2              \\
TextLessRAG\dag  &    &\cellcolor{yellow!30}87.3    &84.0   &80.6    &81.8     &43.2  &\cellcolor{yellow!30}92.6     &61.3     \\
\bottomrule
\end{tabular}
\caption{QA result measured by GPT-4o. For each model, we evaluated the QA performance under two input settings: top-5 retrieved images and gold page marked with ``\dag". Asterisk ``*"  indicates the top-5 results were further refined by layout reranking. We highlight the best results obtained with gold and top-5 page inputs in \sethlcolor{yellow!30}\hl{yellow} and \sethlcolor{green!30}\hl{green}, respectively.}
\label{table:qa}
\end{table*}

We evaluate our approach along three dimensions: \textbf{Retrieval}, \textbf{QA}, and \textbf{Latency}. Retrieval is measured using nDCG@5, while QA accuracy is assessed based on GPT-4o’s averaged judgments. Latency is benchmarked on a single 80GB A100 GPU to ensure consistency.
% Further details of the experimental settings are provided in the repository.

\subsection{Retrival Result}
Table \ref{table:retrive} reports \textbf{Retrieval} results, comparing three text-only methods—BM25\cite{BM25}, E5\cite{(e5}, and NV-Embed-v2\cite{nvembeding}— against five visual models(with ground truth text input as query) — CLIP\cite{clip}, DSE\cite{dse}, VisRAG-Ret\cite{visrag}, VDocRAG\cite{vdocrag}, and ViDoRAG\cite{vidoragvidoseek}—as baselines. For text-based retrieval, OCR text was  extracted using Tesseract\footnote{\url{https://github.com/tesseract-ocr/tesseract}}.
Retrieval experiments were then conducted independently within each retrieval pool across seven datasets.

Our proposed method, TextLessRAG, which adopts a novel speech-based retrieval paradigm, demonstrates competitive performance. 
Compared with three text-to-text retrieval approaches, it consistently and substantially outperforms all of them. 
This indicates that storing the knowledge base as image embeddings directly encoded by a vision encoder yields stronger representational capacity. 
Such an approach not only eliminates the need for OCR but also facilitates query retrieval over multimodal content such as tables and charts.

When compared with text-to-image retrieval methods, TextLessRAG—although not surpassing the SOTA ViDoRAG overall—outperforms the other four baselines and achieves the best results on dataset Vidoseek. 
Moreover, on ChartQA, SlideVQA, MMlong and CDR, the performance gap with SOTA remains within 2\%.
These findings highlight the significant potential of replacing text with speech for image-based knowledge retrieval. 
Furthermore, when textual queries are converted into speech, the loss gap from TTS in retrieval accuracy seemed minimal, enabling direct retrieval with queries spoken in diverse voices while maintaining performance comparable to SOTA models.

Based on the model performance consistency across datasets, we observe that our CDR dataset is easier than Vidoseek but more challenging than MMLong. 
Notably, even with the retrieval pool expanded fivefold(from 5k to 30k), the TextLessRAG retrieval method maintained an accuracy above 87, underscoring its strong capability.
% 对与纯文本检索，我们预先使用Tesseract提取OCR文本
% 我们独立的在各自的检索池使用同样的query在七个数据集上进行检索实验。
% 可以看到ttt使用全新的语音检索方式，其检索效果是具有竞争力的。
% 对比使用文本检索文本的三种方式我们的ttt一致的大幅超过了他们全部。
% 可以看出使用视觉编码器直接将知识库作为图像编码存储embeding的表达能力更强，这省去了OCR步骤的同时更有助于query检索表格图表等多模态内容。
% 对比使用文本检索图像的方式，ttt虽然整体不及SOTA模型xxx，但是它超过了其余四个方法，并且在mmm上拿到了第一。
% 而且在XXX， SSS 和 PPP 中 与SOTA的差距是小于百分之2的
% 这说明使用语音代替文本来检索图片知识库的方式具备巨大潜力。
% 同样的文本在转化为语音后查询知识库后带来的检索精度损失微小，直接将各种音色的语音query直接查询，是能和sota模型达到一样的水准。

% 然后根据模型的一致性表现可以看到我们的DDD数据集比aaa简单比ppp难。
% 但值得注意的是，在检索池扩大至五倍的条件下，多模态检索方案xxx依旧达到了87以上，可以看出其能力强。
\subsection{QA Generation Result}
% Table \ref{table:qa}
% 表x展示了端到端的问答回答质量结果。
% 与检索类似，整体上xxx虽然表现出次优效果但是大幅度超过其余的纯文本方法以及多模态方法。
% 在top5输入情况xxx 在 xxx 到了最好，
% 在使用 layout reranking的情况下，在xxx 和 xxx 达到了最好
% 在 使用 ground truth 输入情况下，xxx和xxx上也到了最好
% 可以看到全流程下来xxx的RAG以及回答上限的表现并不比sota模型xxx逊色
% 这同时反映了在问答阶段使用语音口语输入的巨大潜力
% 而layout reranking ，依据语音筛选出更贴合细粒度的图片内容，使这潜力以及回答的质量得到了进一步的发挥和提升

% 虽然xxx没有在所有数据集上表现的最好，但是也大幅度超过纯文本方法，以及部分数据集上达到SOTA，如xxx的11.2和xxx的12.1
% 考察golden label 页面的输入设置，它体现了当前generator 的回答上界，可以发现除却xx和xx数据集，xxx的表现均略低于yyy。
% 这不得不承认，将文本输入转化为语音，突出方便的同时确实带来了小部分性能折损。
% 但是对比其余的模型，xxx却远远超过了xxx和xxx。
Table \ref{table:qa} presents the end-to-end QA performance results. 
Although TextLessRAG does not achieve the best performance across all datasets, it substantially outperforms purely text-based methods and even reaches state-of-the-art results on certain datasets, such as 87.3 on ChartQA and 93.4 on Vidoseek.
Using golden-label pages as the input, which represent the upper bound of the generator, we find that it is slightly inferior to ViDoRAG on most datasets. Although converting text to speech introduces a minor performance drop, TextLessRAG still substantially outperforms the rest models.
With top-5 inputs, it also achieves the best performance on InfoVQA of 79.4 and SlideVQA of 82.6. 
% When incorporating layout reranking, it further attains the best results on xxx and xxx. 
% Under ground-truth inputs, xxx also achieves the best performance on xxx and xxx.

We then examine the effect of reranking and observe consistent improvements in the answers across all datasets. Notably, on the InfoVQA and SlideVQA datasets, reranking reverses the advantage of the gold setting, with the performance gains from top-5 inputs leading to superior results. Remarkably, even on the Vidoseek, where retrieval quality is already strong, reranking enables the top-5 answers to surpass those obtained with gold inputs.

These findings indicate that across the entire pipeline, the RAG and QA upper bound performance of TextLessRAG is comparable to that of the SOTA model. 
This further underscores the great potential of using spoken queries in the QA stage. 
Moreover, layout reranking, which leverages speech to select image content with finer granularity, enhances both the potential and the overall quality of answers.

\subsection{Accuracy and Latency Analysis}
\begin{figure}[t] % 这里控制图片位置：h=here, t=top, b=bottom, p=page
    \centering
    \includegraphics[width=0.5\textwidth]{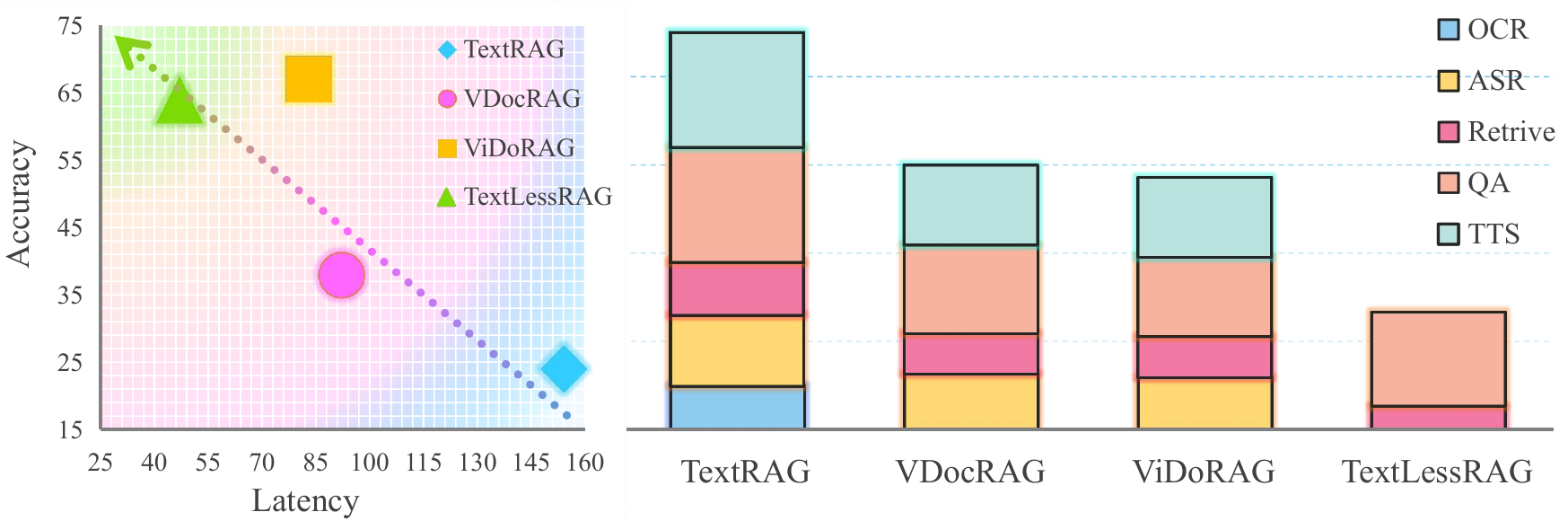} % 图片路径和宽度
    \caption{End-to-end accuracy and latency analysis.}
    \label{fig:latency} % 标签，用于引用
\end{figure}
% 图x综合评估了端到端RAG的回答质量和耗时。
% 我们在评估回答正确率基础上额外计算给出答案的平均耗时。
% 在a图里我们结合端到端回答时延和回答准确率来评估整个pipeline。
% 可以发现xxx兼顾了回答效果的效率.
% 相比于xxx损失微少正确率的情况下回答速率得到了大幅度提升
% 可以看到其次是VdocRAG，他比ViDocRAG 稍微慢一点但是正确率却急速下降，最后是纯文本方法，又慢又易错

% 在图b里我们拆开各个部分耗时，探究慢的原因。
% 可以发现除去通讯消耗，ASR(非流式) TTS(by大模型生成) 和 OCR 依次占据了大量时间。
% 可以看出我们xxx的端到端设计同时直接将语音编码用于检索和生成，直接对视觉特征检索，直接结合视觉特征生成，中间过程实现了全流程去文本化。
% 这种去文本化的设计一方面省去了额外的计算成本以及通讯成本，另一方面也规避了级联的阶段性误差传递，使得模型生成更准确和更快速的响应。

% 然后考察reranking的作用，发现经过reranking筛选后的回答在所有数据集都一致的提高了。
% 特别是在xx和xx数据集上，将原本gold设置的优势反转，在经过rerank后，top5的效果增长促使性能的反超。
% 甚至在检索效果好的yyy数据集上， 使用rerank后，top5的回答效果超过了gold。

Figure\ref{fig:latency} provides an evaluation of end-to-end RAG in terms of both answer quality and latency. 
On the left, we jointly evaluate the end-to-end response latency and answer accuracy to assess the overall pipeline. 
We observe that TextLessRAG strikes a favorable balance between effectiveness and efficiency: compared with ViDoRAG, it achieves a substantial speedup while incurring only a negligible loss in accuracy.
VdocRAG is slightly slower than ViDocRAG but exhibits a steep drop in accuracy, while the pure text method is both slower and more error-prone.

In Figure\ref{fig:latency}(right), we break down the latency of each component to identify bottlenecks. 
Excluding communication overhead, we find that ASR, TTS, and OCR contribute the largest share of latency. 
By contrast, our textless design leverages direct speech encoding for both retrieval and generation, operating directly on visual features for retrieval and response generation. 
This end-to-end de-textualized pipeline not only eliminates additional computational and communication costs but also avoids error propagation across cascaded stages, ultimately enabling more accurate and faster responses.

\section{Conclusion}
% 本工作里，我们首次搭建了一个基于语音输入和语音输出的视觉文档知识库RAG pipeline TextLessRAG。
% 同时我们也首次构建了一个相关的双语测评基准，并标注了首个开源的中文的视觉文档RAG数据集。
% 最终实验证明我们的textless design 省去了OCR，ASR，TTS 的时候节约了大量的推理时间，
% 并且在layout rerank 策略的帮助下保证了，快速检索回答的细粒度block级别感知和高正确率答案生成。
% 总的来说，这次首次对基于语音的视觉RAG的探索挖掘出了语音知识库问答的巨大潜力，相信本工作将有助于多模态大语言模型日后更丰富更便捷的场景落地
In this work, we present TextLessRAG, the first RAG pipeline for visual document knowledge bases with speech-based input and output. 
We also introduce the first bilingual benchmark for this task and release the first open-source Chinese visual document RAG dataset. Experimental results demonstrate that our textless design eliminates the need for OCR, ASR, and TTS, substantially reducing inference time. 
Moreover, with the aid of a layout-based reranking strategy, our approach ensures fine-grained block-level awareness in retrieval and generates highly accurate answers. 
Overall, this pioneering exploration of speech-driven visual RAG reveals the significant potential of speech-based knowledge base question answering and we believe it will facilitate broader and more practical applications of multimodal large language models.

\vfill\pagebreak

% \section{REFERENCES}
% \label{sec:refs}
% \cite{Qwen2.5-Omni}
% List and number all bibliographical references at the end of the
% paper. The references can be numbered in alphabetic order or in
% order of appearance in the document. When referring to them in
% the text, type the corresponding reference number in square
% brackets as shown at the end of this sentence \cite{C2}. An
% additional final page (the fifth page, in most cases) is
% allowed, but must contain only references to the prior
% literature.

% Please follow the IEEE Citation Guidelines, \url{https://ieee-dataport.org/sites/default/files/analysis/27/IEEE\%20Citation\%20Guidelines.pdf} for formatting of references.

\bibliographystyle{IEEEbib}
\bibliography{refs}

\end{document}